\pdfoutput=1

\documentclass[11pt]{article}

\usepackage{emnlp2021}

\usepackage{times}
\usepackage{latexsym}
\usepackage{amsmath}
\usepackage{amssymb}
\usepackage{bm}
\usepackage{float}
\usepackage{mathtools}
\usepackage{booktabs}
\usepackage{caption} 
\DeclarePairedDelimiter{\norm}{\lVert}{\rVert}
\newcommand\wmd{\operatorname{WMD}}
\newcommand\rwmd{\operatorname{R-WMD}}
\newcommand\lcrwmd{\operatorname{LC-RWMD}}
\newcommand\cmd{\operatorname{CMD}}
\newcommand\wmdec{\operatorname{WMDecompose}}

\usepackage[T1]{fontenc}

\usepackage[utf8]{inputenc}

\usepackage{microtype}

\title{WMDecompose: A Framework for Leveraging the Interpretable Properties of Word Mover's Distance in Sociocultural Analysis}

\author{Author 1 \\ Address line \\  ... \\ Address line
        \And  ... \And
        Author n \\ Address line \\ ... \\ Address line}

\author{Mikael Brunila \\
  McGill University\\
  Department of Geography\\
  \texttt{mikael.brunila@gmail.com} \\\And
  Jack LaViolette \\
  Columbia University \\
  Department of Sociology \\
  \texttt{jack.laviolette@columbia.edu} \\}

\begin{document}
\maketitle
\begin{abstract}

Despite the increasing popularity of NLP in the humanities and social sciences, advances in model performance and complexity have been accompanied by concerns about interpretability and explanatory power for sociocultural analysis. One popular model that balances complexity and legibility is Word Mover's Distance ($\wmd$). Ostensibly adapted for its interpretability, $\wmd$ has nonetheless been used and further developed in ways which frequently discard its most interpretable aspect: namely, the word-level distances required for translating a set of words into another set of words. To address this apparent gap, we introduce $\wmdec$: a model and Python library that 1) decomposes document-level distances into their constituent word-level distances, and 2) subsequently clusters words to induce thematic elements, such that useful lexical information is retained and summarized for analysis. To illustrate its potential in a social scientific context, we apply it to a longitudinal social media corpus to explore the interrelationship between conspiracy theories and conservative American discourses. Finally, because of the full $\wmd$ model's high time-complexity, we additionally suggest a method of sampling document pairs from large datasets in a reproducible way, with tight bounds that prevent extrapolation of unreliable results due to poor sampling practices.

\end{abstract}

\section{Introduction: The Paradox of Word Mover's Distance}

The present paper introduces $\wmdec$, an iteration of the Word Mover's Distance ($\wmd$) \cite{kusner_word_2015} model commonly used for determining the semantic distances between pairs of documents. Leveraging word vectors from models such as word2vec \cite{mikolov_efficient_2013}, fastText \cite{bojanowski_enriching_2016} and GloVe \cite{pennington_glove_2014}, $\wmd$ was presented as a method that was not only hyper-parameter free and thus easy to use, but also highly interpretable \cite{kusner_word_2015}. Arguably, this still makes the model a viable alternative for document similarity tasks, despite the recent and rapid rise of contextual and rich embeddings from Transformer-type models \cite{vaswani_attention_2017} such as BERT \cite{devlin_bert_2019}. However, $\wmd$ is computationally expensive, with the full model running at cubic time complexity \cite{kusner_word_2015}. This may explain why, despite its initial pitch as an interpretable alternative, $\wmd$ has mainly been further developed and applied in ways that focus on decreasing the model's high runtime, while ignoring or undermining the inherent interpretability of the model \cite{atasu_linear-complexity_2017,werner_speeding_2020}.

To our knowledge, no current $\wmd$ implementation provides an out-of-the-box means of retaining word-level information, despite its utility to many research agendas. To confront this paradoxical situation, we introduce a set of methods and Python code for retaining the individual word distances that make $\wmd$ interpretable while simultaneously suggesting a simple trick for efficiently estimating the distance between large sets of documents. Specifically, we propose using implementation of the ``relaxed'' $\wmd$ \cite{kusner_word_2015} with linear time complexity \cite{atasu_linear-complexity_2017} to first estimate the distances between a full set of documents, then using optimal pairing with the Gale-Shapley algorithm \cite{gale_college_1962} on this estimate to find the pairs between two sets of documents that minimize the average pairwise distance. Next, full $\wmd$ is calculated between each pair, while progressively adding the contributions of individual words to the overall distance between document sets. Words are, in turn, grouped using K-means clustering and vector dimensionality reduction to decompose the distances not only by word, but by thematic cluster. Each cluster is defined by its constituent words, and hence highly interpretable. 

Having introduced $\wmdec$, we demonstrate its utility in a social scientific context by exploring interrelated trends over time between two social media corpora: \verb|r/conspiracy| and \verb|r/The_Donald|, the primary Reddit communities for self-identified conspiracy theorists and Donald Trump supporters, respectively. While this presents only a cursory and exploratory engagement with these thematically complex data, we hope that it demonstrates the analytic potential of $\wmdec$ for social science and humanities research.

Finally, we also provide a complementary analysis using a well-known Yelp review dataset, to act as a sanity check on the validity of our method. This analysis can be found in Appendix B.

\section{Previous Work}

\subsection{Word Mover's Distance}

Originally proposed by \citet{kusner_word_2015}, $\wmd$ has become a popular metric of document semantic distance in computational linguistics and related subfields. An innovation and special instance of Earth Mover's Distance \cite{rubner_metric_1998,pele_linear_2008,pele_fast_2009}, $\wmd$ leverages word embeddings to compute the minimum distance required to ``move'' the words from one document to another, providing a measure of document-level semantic (dis)similarity as a sum of the distance required to move individual words from one document to another (see Figure \ref{fig:kusner2}). Since its introduction, many related algorithms and analytic approaches have been proposed for language engineering tasks \cite[e.g.][]{atasu_linear-complexity_2017,huang_supervised_2016,ren_emotion_2018}. More recently, $\wmd$ and its many variations have also been applied to socioculural analyses of data ranging from survey response data \cite{taylor_concept_2020}, to Ancient Greek literature \cite{pockelmann_fast_2020}, to dyadic conversational dynamics \cite{nasir_modeling_2019}.

$\wmd$ is typically parameterized in the following manner \cite{kusner_word_2015, huang_supervised_2016}, using words represented as embeddings produced with algorithms such as word2vec \cite{mikolov_efficient_2013}. Let $x_i \in \mathbb{R}^{d}$ be the $i$th embedding in $d$-dimensional space, drawn from a word embedding matrix $X \in \mathbb{R}^{d \times n}$ representing a vocabulary of $n$ words. Let $\boldsymbol{\mathrm{d}}^a$ and $\boldsymbol{\mathrm{d}}^b$ be two $n$-dimensional, normalized bag-of-words (nBOW) vectors for a pair of documents where $d^a_i$ is the normalized number of times word $i$ occurs in vector $d^a$. $\wmd$ then attempts to find a transportation matrix $\boldsymbol{\mathrm{T}} \in \mathbb{R}^{n \times n}$ that minimizes the total distance required to move all words in the first document to the second document, where $\boldsymbol{\mathrm{T}}_{i,j}$ describes how much of the normalized word vector $d_i^a$ should be transported to the normalized word vector $d_j^b$. Formally, $\wmd$ returns the minimum distance to move from document $\boldsymbol{\mathrm{d}}^a$ to document $\boldsymbol{\mathrm{d}}^b$, given by summing the product of the optimal ``flow'' $\boldsymbol{\mathrm{T}}_{i,j}$ from all words in the two documents with the ``cost'' $c(i,j)$ of moving between each word vector in the two documents:

\begin{align}
\begin{split}
\wmd(\boldsymbol{\mathrm{d}}^a, \boldsymbol{\mathrm{d}}^b) = \underset{\boldsymbol{\mathrm{T}}\geq 0}{\mathrm{min}}\sum^{n}_{i,j=1}\boldsymbol{\mathrm{T}}_{i,j}c(i,j)\\ 
\end{split}
\end{align}

Furthermore, the equation is subject to the constraint that the entirety of the ``mass'' of $\boldsymbol{\mathrm{d}}^a$ should be distributed in the flows to $\boldsymbol{\mathrm{d}}^b$ and vice versa: 
\begin{align}
\begin{split}
\sum^{n}_{j=1}\boldsymbol{\mathrm{T}}_{i,j} = d_i^a &\;\;\forall i \in \{1,...,n\} \\
\sum^{n}_{i=1}\boldsymbol{\mathrm{T}}_{i,j} = d_j^b &\;\;\forall j \in \{1,...,n\}
\end{split}
\end{align}

Even though the original implementation of the $\wmd$ uses Euclidean distance  for the metric $c(i,j)$, the similarity between word embeddings in general and with WMD in particular \cite{yokoi_word_2020} is better captured using the cosine distance\footnote{The use of the word ``distance'' here can be slightly confusing, as it is not quite the same as the ``distance'' in Word Mover's \emph{Distance}. For the latter, the distance between two documents is composed of the cosine distance \emph{and} the ``mass'' of the nBOW representation of the documents to be moved to one another.}:
\begin{align}
\begin{split}
c(i,j)=1 - \frac{\boldsymbol{\mathrm{x}}_i \cdot \boldsymbol{\mathrm{x}}_j}{\norm{\boldsymbol{\mathrm{x}}_i}\norm{\boldsymbol{\mathrm{x}}_j}}
\end{split}
\end{align}

Furthermore, $\boldsymbol{\mathrm{d}}^a$ and $\boldsymbol{\mathrm{d}}^b$ can be normalized using other techniques, such as Term frequency-Inverse document frequency (Tf-Idf), combined with L1-normalization, which allocate more mass to words that are more common in a specific document than the overall vocabulary.

While there is a well-established literature on solving this linear program using the EMD algorithm, doing so is computationally prohibitive. \citet{kusner_word_2015} note that ``the best average time complexity of solving the $\wmd$ optimization problem scales'' $O(V^3 \log V)$, with $V$ denoting the number of unique words in the corpus. Consequently, calculating the distances for large datasets will often prove insurmountable for $\wmd$. \citet{kusner_word_2015} suggest working around this issue by calculating an approximation of $\wmd$, where the distance from the two pairs are calculated with relaxed constraints, so that $\boldsymbol{\mathrm{T}}_{i,j}$ must contain the mass or flow from the source document, but not the target. This ``relaxed'' $\wmd$ or $\rwmd$ is then instead parameterized with only one constraint: 

\begin{align}
\begin{split}
\rwmd(\boldsymbol{\mathrm{d}}^a, \boldsymbol{\mathrm{d}}^b) = \underset{\boldsymbol{\mathrm{T}}\geq 0}{\mathrm{min}}\sum^{n}_{i,j=1}\boldsymbol{\mathrm{T}}_{i,j}c(i,j) \\
\text{subject  to:}
\sum^{n}_{j=1}\boldsymbol{\mathrm{T}}_{i,j} = d_i \;\;\forall i \in \{1,...,n\}
\end{split}
\end{align}

Repeating the process, so that $\rwmd$ is instead calculated from $\boldsymbol{\mathrm{d}}^b$ to $\boldsymbol{\mathrm{d}}^a$, gives us an estimate of the bounds within which the full $\wmd$ must be located. While $\rwmd$ gives only an estimate of $\wmd$ within certain bounds, it has been shown to be a good approximation of the full $\wmd$ \cite{kusner_word_2015}.  Notably, instead of the cubic time complexity of $\wmd$, $\rwmd$ can be performed with quadratic time complexity $O(V^2)$. Furthermore, \citet{atasu_linear-complexity_2017} have demonstrated how to calculate the $\rwmd$ so that the time complexity is reduced from quadratic to linear with the Linear Complexity Relaxed WMD ($\lcrwmd$) algorithm. However, this solution does \textit{not} retain the distances contributed by individual words to the $\rwmd$.

\subsection{WMD and interpretability}

\begin{figure}[t]
    \centering
    \includegraphics[width=0.5\textwidth]{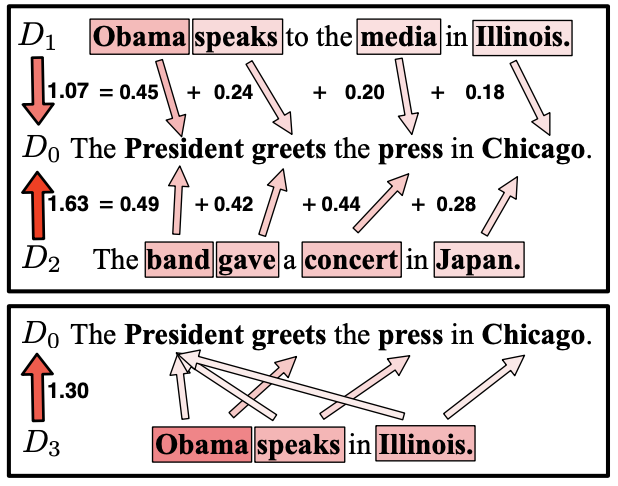}
    \caption{An illustration of $\wmd$ in action by \citet{kusner_word_2015}. (Top:) The components of the $\wmd$ metric between a query $D_0$ and two sentences $D_1$ , $D_2$ (with equal BOW distance). The arrows represent flow between two words and are labeled with their distance contribution. (Bottom:) The flow between two sentences $D_3$ and $D_0$ with different numbers of words. This mismatch causes the $\wmd$ to move words to multiple similar words.}
    \label{fig:kusner2}
\vspace{-10pt}
\end{figure}

While these and other suggested tricks and improvements
\cite[e.g.][]{tithi_efficient_2021-1,werner_speeding_2020,yokoi_word_2020} for efficient $\wmd$ make the algorithm a feasible and powerful tool for comparing the semantic distance between large sets of documents, they have been introduced with little regard to the algorithm's initial claims to intuitive and interpretable explanations for document (dis)similarities. Consider, for example, Figure \ref{fig:kusner2}, introduced by \citet{kusner_word_2015} as an example of how the distances between three documents could be decomposed into different parts. Indeed, for sophisticated NLP techniques such as $\wmd$ to be maximally useful for sociocultural analysis, the possibility to decompose document-level results into interpretable lexical information is key. For example, if the mean $\wmd$ document distances between two longitudinal corpora sampled at some time $t_0$ and again at some later time $t_1$ shrink, it might indicate that the corpora have become more similar. However, the change in distance by itself would tell the curious analyst little about the particular lexical phenomena responsible for the semantic changes, and (crucially) how these might relate to extralinguistic social and symbolic processes. In order to tell the full story, fluctuations in distance must be decomposed into interpretable parts.

To invoke Danilevsky et al.'s (\citeyear{danilevsky_survey_2020}) classification scheme of explanation in NLP, $\wmd$ provides explanations that are \textit{local} (i.e. the distance between any pair of documents can be decomposed to individual words) and \textit{self-explanatory} (i.e., no \textit{post hoc} processing of the model outputs is necessary). However, these explanations have, to the best of our knowledge, not been leveraged in applied research with $\wmd$, most likely due to the prohibitive computational cost of calculating the full $\wmd$ and the neglect of interpretability in more efficient elaborations of the model. Moreover, applied $\wmd$ research has to date generally begun with an \textit{a priori} interest in the relationship between documents and particular ``concept'' words \cite[e.g.]{stoltz_concept_2019} or predefined topics \cite[e.g.]{wu_topic_2017}. Although such analyses are perfectly valid, a fully inductive relationship to words of interest can be useful when the analyst does not have or desire strong assumptions about what lexical changes underlie the phenomena of interest.

To these ends of inductive interpretability, we propose what we believe to be a novel analytic pipeline which combines $\lcrwmd$ decomposed at the word level, optional document pairing based on the Gale-Shapley matching algorithm to ensure robustness, and t-SNE \cite{maaten_visualizing_2008} reduced word vectors with K-means clustering \cite{Lloyd1982LeastSquaresQuantization, Elkan2003UsingTriangleInequality} to enhance interpretability by inducing higher-level thematic groupings. To assist future researchers, we additionally provide a Python package and example code notebooks available on Github.\footnote{\url{https://github.com/maybemkl/wmdecompose}}

\section{Data}

\subsection{The corpus: conspiracy theories and American conservatism online}

Several researchers have identified conspiratorial thinking as a feature of the (post-)Trump era of American conservative politics
\cite{bracewell_gender_2021,hellinger_conspiracies_2018,polletta_deep_2019}. Many such studies have been theoretical and/or qualitative \cite[e.g.][]{barkun_president_2017,stecula_how_2021}, survey-based \cite[e.g.][]{federico_role_2018,miller_conspiracy_2016,uscinski_why_2020}, focused on patterns of ``misinformation'' dissemination on social media \cite[e.g.][]{benkler_study_2017,marwick_media_nodate}, or on the discourse of political elites \cite[e.g.,][]{hornsey_donald_2020,neville-shepard_post-presumption_2019}. The present paper chooses a different, though not unprecedented, approach of studying large user-generated text corpora from \url{reddit.com}. Specifically, we analyze ``self posts''\footnote{Self posts are forum submissions which contain only text as opposed to links to external sites. We use self posts in lieu of comments, as the latter tend to be shorter and less orderly due to their nested structure. Further, self posts tend to be more regulated by forum moderators, suggesting that they are more likely to reflect the norms of the community.} from the \verb|r/The_Donald| and \verb|r/conspiracy| communities (``\textit{subreddits}''); the former was the main subreddit for Donald Trump supporters before being banned in June 2020, while the latter remains the largest subreddit for self-identified conspiracy theorists. Others have examined both \verb|r/The_Donald| and its role in alt-right politics \cite{massachs_roots_2020,ribeiro_does_2020,shepherd_gaming_2020}, as well as conspiracy theorists on the platform \cite{klein_pathways_2019,phadke_what_2021,samory_conspiracies_2018}. Research has indicated a sizable and statistically significant overlap in users between the two subreddits \cite{massachs_roots_2020,nithyanand_online_2017}. However, these observations have been largely based on subreddit co-subscriber networks, and have paid comparatively less attention to large-scale linguistic patterns over time, a gap to which we hope to contribute.

Data was collected using the Pushshift Reddit dataset publicly available on Google BigQuery \cite{baumgartner_pushshift_2020}. Because we are interested in change over time, we delineate two discontinuous periods of interest: $t_0$, defined as the twelve months following the creation of \verb|r/The_Donald| (July 11, 2015--July 11, 2016), and $t_1$, the final twelve months available in the dataset (August 31, 2018--August 31, 2019). These two year-long snapshots, separated by roughly two years, offer a glimpse of the relationship between conspiratorial language and the discourse of self-identified Donald Trump supporters during the Trump presidency.

\subsection{Sampling and preprocessing}

To reduce computation and run-time, we sample 5,000 posts from each subreddit at $t_0$ and $t_1$ for a total of 20,000 posts. Random sampling was restricted to those posts at least 30 words long (to ensure that documents contain adequate lexical information), and having a positive score (2 or greater, as Reddit posts start with a score of 1) to ensure that the lexical content is generally representative of the community. Once sampled, the text is preprocessed using a standard pipeline for NLP applications. The specifics of this pipeline are described in Appendix A. The processed dataset contains 1,509,553 tokens, 596,596 tokens from \verb|r/The_Donald| and 993,957 from \verb|r/conspiracy|.

\section{Methods}
\subsection{Embedding and clustering}

After preprocessing, words were converted to vectors using a fine-tuned word2vec model introduced in \cite{mikolov_efficient_2013}, originally pre-trained on Google News Vectors containing about 300 billion words. Fine-tuning details and hyperparameters are included in Appendix A. Due to the large number of unique words, and to increase the interpretability of the final outputs, words were clustered on the basis of their embedding vectors using t-SNE, an algorithm that is commonly used to reduce the dimensionality of word embeddings while still preserving the original structure of the higher-dimensional form \cite[for uses with WMD, see:][]{huang_supervised_2016, gulle_topic_2020}. In our case, we reduce the dimensionality of the original word vectors from 300 to two and use K-means on the reduced dimensions to generate 100 clusters of words according to their semantic similarity. Clustering words allows us to examine not only the changing usage of individual words across subreddits, but also of these higher-level thematic groupings. The number of clusters was chosen heuristically after inspecting both elbow plots and silhouette scores for clusters in the range of 10 to 200. Our method is robust to using raw embeddings as well as other popular reduction techniques before clustering, such as UMAP \cite{mcinnes_umap_2020}.

\subsection{WMDecompose}

We now introduce $\wmdec$, the core contribution of this paper, which provides the ability to examine the word-level distances required to move between two sets of documents, such as the self posts in \verb|r/conspiracy| and \verb|r/The_Donald|. The comparison of these documents happens, on the one hand, through retaining the word-level distances of moving between pairs of documents from each set and, on the other hand, by clustering words and aggregating their added distances by cluster.

More generally: Given two sets of documents, $S^a$ and $S^b$, the matrix $\boldsymbol{\mathrm{T}}_{i,j}$ of flows between all the individual documents in both sets, and clusters for the input word vectors, $\wmdec$ returns the following information twice, once in terms of movement from the first set to the second and once in terms of movement from the second set to the first:

\renewcommand{\labelenumi}{\Alph{enumi}}
\begin{enumerate}
  \item The aggregate word-level $\wmd$ or $\wmd_w$ for each word $w$ in a vocabulary $V$ when moving all documents from one set of documents $S^a$ to another set  $S^b$.
  \item The aggregate cluster-level or Cluster Mover's Distance ($\cmd_c$) for each cluster $c$ when moving all documents from $S^a$ to $S^b$, with keywords for each $c$ determined by the words with the highest $\wmd_w$ within the cluster.
\end{enumerate}

The $\wmd_w$ gives us a good sense of the most important words separating the two sets. The $\cmd_c$ allows us to organize this information by cluster, with interpretable cluster keywords that are dynamically ranked, depending on their importance for the particular case at hand. 

More formally, $\wmd_w$ is calculated in the following manner. Let $w^a$ and $w^b$ be two words contained in $S^a$ and $S^b$, respectively, and let $\wmd_{w^a}$ be the total distance accumulated by word $w^a$ when moving from $S^a$ to $S^b$. Next, if $n$ is the number of documents in $S^a$ that contain the word $w^a$, $m$ is the number of words in some document $d^b$ that $w^a$ is distributed over,  $\boldsymbol{\mathrm{t}}_{i,j}$ is the vector of flows from $w^a$ in document $d^a_i$ to each word $w^b_j$ in $d^b$, and $c(w^a,w^b_j)$ is the cost in terms of cosine distance to move from each $w^a$ to each $w^b_j$, then

\begin{align}
\begin{split}
\wmd_{w^a} = \sum^n_{i=1}\sum^{m}_{j=1}\boldsymbol{\mathrm{t}}_{i,j} c(w^a,w^b_j)
\end{split}
\end{align}

$\wmd_{w^b}$ is counted in the same manner, only this time using the flow and cost from words in $S^b$ to words in $S^a$. 

The $\cmd_{c^a}$, i.e. the $\cmd_{c}$ for movement from $S^a$ to $S^b$, is then calculated simply by summing over the aggregated distances of all $w^a \in c$, i.e. all words $w^a$ belonging to cluster $c$. If there are $p$ such words in $S^a$ for some cluster $c$, then

\begin{align}
\begin{split}
\cmd_{c^a} = \sum_{k=1}^{p}\wmd_{w^a}\;\;\forall w^a \in c^a\\ 
\end{split}
\end{align}

Again, $\cmd_{c^b}$ is counted in the same manner. Furthermore, the total $\wmd_{S^a}$ when moving all documents from $S^a$ to some pair in $S^b$ can consequently be described as

\begin{align}
\begin{split}
\wmd_{S^a} = \sum \wmd_{w^a} = \sum \cmd_{c^a}
\end{split}
\end{align}

In order to further accentuate the differences between the word-level distances $\wmd_{w^a}$ and $\wmd_{w^b}$, we subtract the $\wmd_{w^a}$ for each word $w^a$ in the vocabulary of $S^a$ from the corresponding $\wmd_{w^b}$ (if it exists) in the other set and vice versa. This way results will not be cluttered by words for which the $\wmd_w$ is very similar across sets and instead the differences between the two sets will be highlighted. Hence, the final $\wmd_{w^a}$ with ``difference'' or $\wmd_{w^a}^d$ for word $w^a$ is

\begin{align}
\begin{split}
\wmd_{w^a}^d = \wmd_{w^a} - \wmd_{w^b}
\end{split}
\end{align}
if there exists such a $w^b$ that $w^b = w^a$. In other cases, the $\wmd_{w^a}^d$ is just equal to the $\wmd_{w^a}$. For the rest of the paper, we will use $\wmd_{w^a}$ as a short-hand when writing about $\wmd_{w^a}^d$ as the model with difference yields far better and more interpretable results and is the only one we will consider in our analysis.

Finally, the $\wmd_{w}$ values can be calculated using $\wmd$ with or without relaxed constraints. However, while the $\lcrwmd$ cannot be used, we will next suggest a way in which its speed can be leveraged to yield reproducible and conservative estimates of the $\wmd_{w}$ values.

\subsection{Gale-Shapley matching for WMD}

To run $\wmd$ on all 50 million pairwise combinations of documents in our corpus ($5000^2$ at both $t_0$ and $t_1$) would unfortunately be computationally prohibitive in many instances. With larger document sets, even using $\rwmd$ with quadratic time complexity could be too costly in terms of time and computational resources, while using the $\lcrwmd$ with linear time complexity sacrifices word-level decomposability. One solution would be to take a random sample from both sets of documents. However, this is not always ideal, as we will demonstrate next.

\begin{figure}
    \centering
    \includegraphics[width=0.4\textwidth]{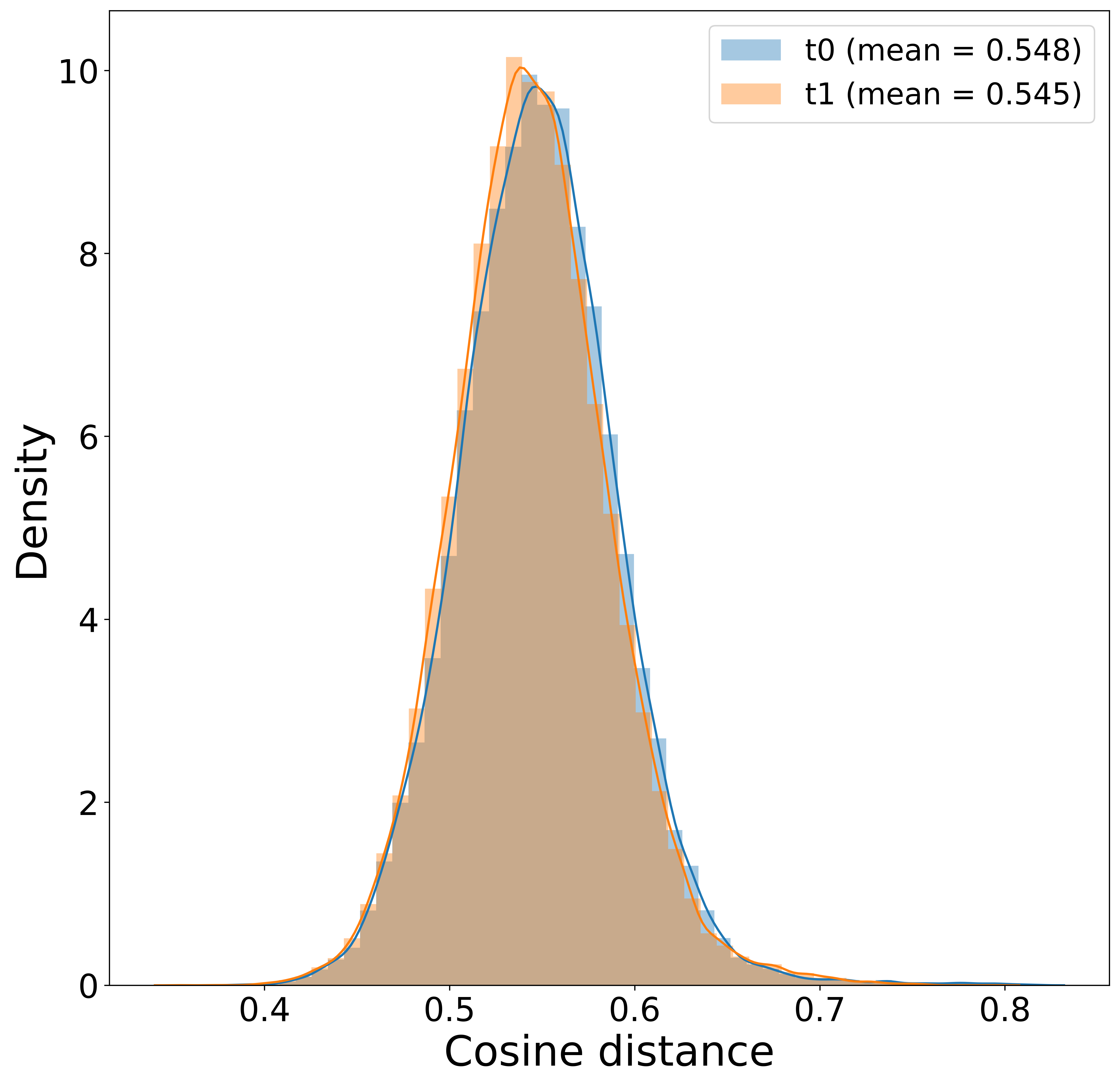}
    \caption{Distribution of $\wmd$ distances for random document pairs (one from each subreddit) at $t_0$ and $t_1$.}
    \label{fig:wmd_hist}
\end{figure}

We first generate 50,000 random pairs of documents (each pair containing one post from each subreddit) at both $t_0$ and $t_1$, yielding a total of 100,000 random pairs. The distribution of distances is displayed in Figure \ref{fig:wmd_hist}. Although a \textit{t}-test confirms high significance (\textit{t} = 13.71), the measured difference in means is very small (0.548 at $t_0$, versus 0.545 at $t_1$), and it is difficult to determine to what meaningful extent this reflects increased similarity between user language in \verb|r/The_Donald| and \verb|r/conspiracy| over time. We will return to this point later, in Section 5.2.

Looking at Figure \ref{fig:wmd_hist}, it is clear that merely drawing a small random sample from both sets could produce unreliable results. Researchers working with very long documents and/or lacking computational resources might be limited to drawing small samples if running the full $\wmd$ model. If asking whether there is any evidence of document-level semantic alignment---i.e., a reduction in mean document pair distance over time---present in our corpus in the first place, using a random sample to offset computational costs could therefore produce misleading results. We therefore introduce a second contribution of $\wmdec$: the option of effective and consistent pairing of documents using the Gale-Shapley (GS) stable matching algorithm \cite{gale_college_1962} to reduce the computational burden incurred by analysis.

Originally introduced through the hypothetical problem of pairing colleges with applicants, the GS algorithm iteratively finds the optimal match between two sets of equal size, given the preferences of all members of the two sets. The optimal match is biased towards the party taking initiative, i.e. the ``suitor'' \cite{Dubins1981MachiavelliGaleShapleyAlgorithm, Iwama2008SurveyStableMarriage}. To find the ``preferences'' of documents, we utilize $\lcrwmd$ to first get an approximation of distances between all pairs from two sets of documents. Each document in a set then ``prefers'' the closest document in the other set. Given these preferences, we can use GS to find the pairs which minimize the distance between the two sets. We posit that GS ensures that our document pairs represent a conservative estimate of the distance between the two sets, making our method reproducible while also ensuring the robustness of results. By reducing computational costs, GS pairing is additionally aligned with mounting calls for NLP and other ML researchers to intentfully pursue algorithms which minimize the growing environmental toll of their technologies \cite{bender_dangers_2021, strubell_energy_2019}.

For our specific case study, we first calculate the $\lcrwmd$ from \verb|r/conspiracy| to \verb|r/The_Donald| and vice versa. This computation on 50M total pairs requires less than an hour on a regular laptop. Next, these relaxed distances from each document $i$ in \verb|r/conspiracy| to all $N$ documents in \verb|r/The_Donald| are used as the preferences of document $i$ vis-a-vis documents ${1,...,N}$. We are primarily interested in movement from \verb|r/conspiracy| to \verb|r/The_Donald|, hence the former is given the role of ``suitor''.

Then, the Gale-Shapley algorithm is executed to find the optimal match, i.e., the set of document pairs that is least costly to move between \verb|r/conspiracy| and \verb|r/The_Donald|. Effectively, this shows one optimal solution to moving the mass of words from former to the latter, if each document in the first set were to be transferred only to the most similar document in the other set. All these steps are repeated for the documents at $t_0$ and $t_1$. To summarize, by matching pairs of documents which are more similar in semantic space, $\wmdec$ produces conservative measures of semantic distance between corpora. Thus, in addition to providing a non-random, reproducible means of reducing computing costs while retaining full $\wmd$ calculations, GS ensures the robustness of findings.

\subsection{Exploring lexical trends with WMDecompose}

Once we have the optimal pairs between \verb|r/conspiracy| and \verb|r/The_Donald|, we can run full $\wmd$ for the two sets of pairs, i.e. the set at $t_0$ and the set at $t_1$\footnote{Relaxed $\wmd$ can also be used to further speed up the calculations. While using $\rwmd$ gives similar results as the full $\wmd$, we focus here on the full $\wmd$ results as this is the baseline we are looking to establish.}. We do this using \verb|wmdecompose|, a Python package written specifically for this paper with EMD executed using the \verb|PyEMD| library under the hood \cite{pele_linear_2008, pele_fast_2009}.\footnote{\url{https://github.com/laszukdawid/PyEMD}}

\section{Results}

\subsection{Overall document distance}

After generating document pairs with the Gale-Shapley algorithm, it is worth asking whether mean document distances using GS pairs follow a similar pattern than those generated by random pairs (as in Figure \ref{fig:wmd_hist}). As it turns out, they do; both are normally distributed, with pair distances having a mean of 0.478 at $t_0$ (versus 0.548 with random pairing), and 0.469 at $t_1$ (versus 0.545 with random pairing). A \textit{t}-test again shows significance (\textit{t} = 8.05), but the absolute difference is small enough as to be difficult to interpret at the document level. We can therefore conclude that, when pairing documents on the basis of semantic proximity with GS and when pairing documents randomly, some modest but statistically significant reduction in distance is taking place from $t_0$ to $t_1$. Without further analysis, however, we cannot conclude much else.

\subsection{Identifying distinguishing words with WMDecompose}

As we have stated throughout, document-level distance measures tell one nothing about the nature and source of that distance. A researcher doing exploratory analysis of a new dataset might look at the relatively small, albeit significant, difference in mean document distance at $t_0$ and $t_1$ and conclude, absent any other contextual information, that further analysis is not warranted. 

\begin{figure}
    \centering
    \includegraphics[width=0.5\textwidth]{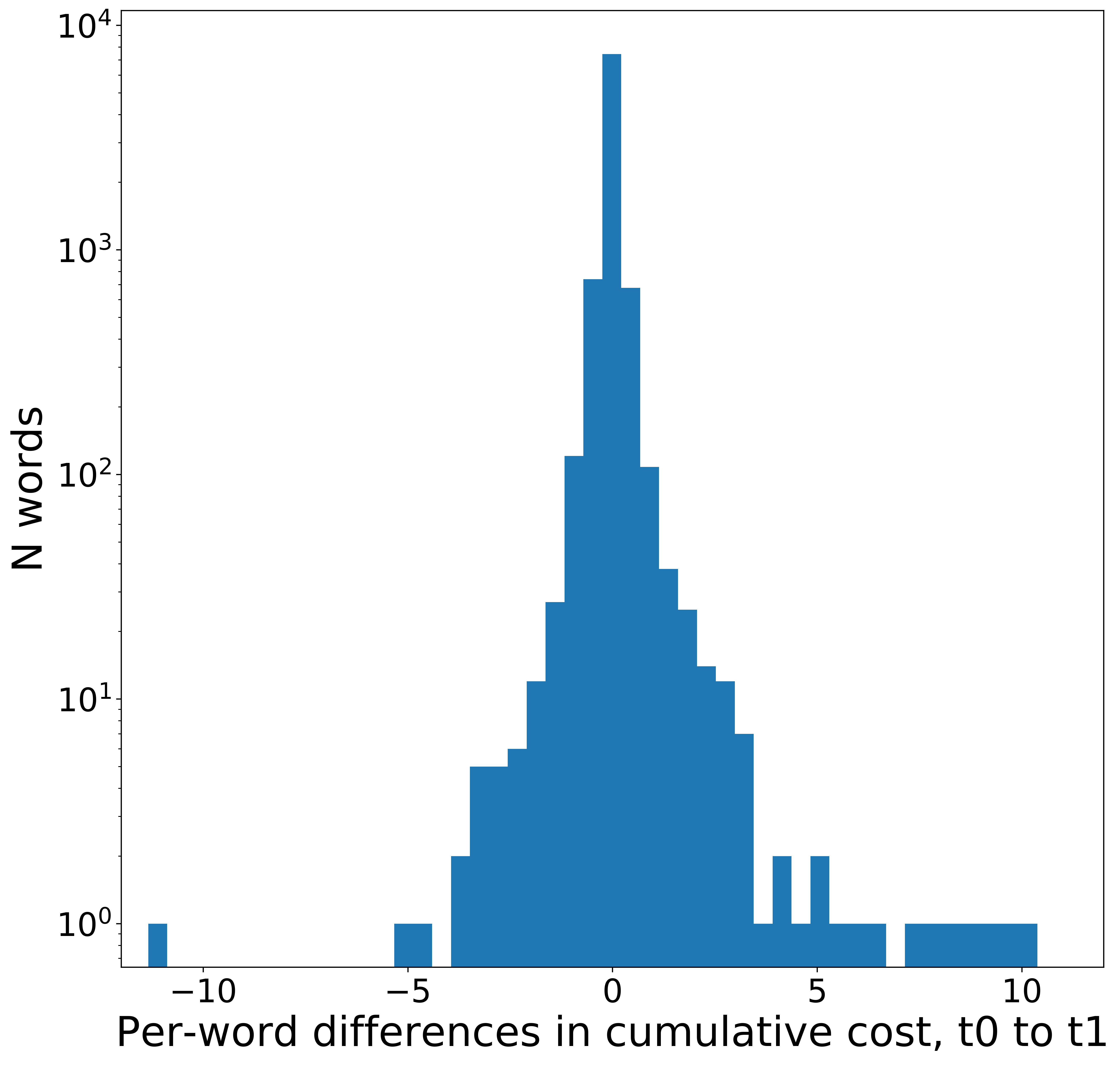}
    \caption{The distribution of word distance changes over time, calculated as the aggregate distance each word contributed to $t_1$ document pairs minus the aggregate distance it added to $t_0$ document pairs. Despite the general stability of document-level distances,  many individual words show considerable differences in distance contributed from $t_0$ to $t_1$.}
    \label{fig:word_cost_diff}
\end{figure}

However, distances between documents merely represent the distances between their aggregate words. As such, it is quite possible that relative document-level stability is masking a much greater degree of variation between $t_0$ and $t_1$ for specific words, and these words might merit further attention. Figure \ref{fig:word_cost_diff} displays the distribution of changes in word cost (i.e., the total distance contributed at $t_1$ minus the total distance contributed at $t_0$) for each unique word in the corpus. As we can see, despite the longitudinal stability of the corpora at the document level, a great deal of lexical and semantic change over time is nevertheless taking place. Though the majority of words show little change in this regard, many words which significantly distinguished the corpora at $t_0$ no longer do at $t_1$, and vice versa. Our suggestion is that these words might be qualitatively instructive, motivating a closer inspection of the lexical data produced by $\wmdec$. We therefore turn to $\wmdec$ to examine the sorted list of words and clusters which most distinguish our corpora at each time period, displayed in Table \ref{tab:subreddit_words}.

These top words conform nicely with expectations, with vocabulary directly associated with party and electoral politics characterizing \verb|r/The_Donald|, and words associated with conspiracy theories characterizing their eponymous subreddit. Further, we see the clear effects of temporality and current events: words associated with the 2016 election cycle (\textit{bernie, hillary, cruz, delegate}) disappear at $t_1$, while more general political terms (\textit{democrat, president, liberal}) appear; for \verb|r/conspiracy|, \textit{epstein} and \textit{cia} enter the top ten most distinctive words, while \textit{isis} drops out. 

\begin{table}[t!]
\centering
\begin{tabular}{l l l l}
\toprule
\begin{tabular}{l c}
    \multicolumn{2}{c}{The\_Donald ($t_0$)} \\
    \textbf{Word} & \textbf{Cost}\\
    \midrule
    trump & 45.85 \\
    vote & 14.96 \\
    bernie & 10.79 \\ 
    hillary & 10.49 \\ 
    cruz & 10.01 \\ 
    win & 9.58 \\ 
    donald & 9.58 \\
    donald\_trump & 8.81 \\
    delegate & 7.95 \\ 
    cuck & 7.50 \\ 
  
    \end{tabular}
    \begin{tabular}{lc}
    \multicolumn{2}{c}{conspiracy ($t_0$)} \\
    \textbf{Word} & \textbf{Cost}\\
    \midrule
    conspiracy & 8.79 \\
    government & 6.67 \\
    world & 3.90 \\ 
    israel & 3.72 \\ 
    find & 3.66 \\ 
    video & 3.56 \\ 
    information & 3.08 \\ 
    documentary & 3.00 \\ 
    theory & 2.97 \\ 
    isis & 2.97 \\ 
    
    \end{tabular}
 \\
 \bottomrule
 \end{tabular}
 
 \begin{tabular}{l l l l}
 \begin{tabular}{l c}
    \multicolumn{2}{c}{The\_Donald ($t_1$)} \\
    \textbf{Word} & \textbf{Cost}\\
    \midrule
    trump & 12.83 \\
    vote & 10.70 \\
    democrat & 6.50 \\ 
    president & 6.45 \\ 
    dem & 5.90 \\ 
    left & 5.12 \\ 
    pedes & 4.76 \\ 
    election & 4.74 \\ 
    liberal & 4.70 \\ 
    win & 4.33 \\
    
    \end{tabular}
    \begin{tabular}{lc}
    \multicolumn{2}{c}{conspiracy ($t_1$)} \\
    \textbf{Word} & \textbf{Cost}\\
    \midrule
    conspiracy & 11.17 \\ 
    epstein & 8.97 \\  
    video  & 5.30 \\ 
    control  & 4.97 \\ 
    vaccine  & 4.46 \\ 
    theory  & 3.94 \\ 
    world  & 3.69 \\ 
    find  & 3.46 \\ 
    government  & 3.33 \\ 
    cia  & 2.89 \\ 

    \end{tabular}
 \\
 \bottomrule
 \end{tabular}
\caption{Words contributing the most cost between subreddits at $t_0$ and $t_1$.}
\label{tab:subreddit_words}
\vspace{-10pt}
\end{table}

\begin{table*}[]
    \centering
    \resizebox{\textwidth}{!}{%
    \begin{tabular}{l c | c c c c c c c c}
    \toprule
    Subcorpus & Cluster Theme & Word 1 & Word 2 & Word 3 & Word 4 & Word 5 & Word 6 & Word 7 & Word 8\\
    \midrule
     \textbf{The\_Donald ($t_0$)} & \textbf{2016 primary election} & cruz & kasich & campaign & ted\_cruz & republican & rubio & party & vp \\ \toprule
     \textbf{The\_Donald ($t_0$)} & \textbf{pro-Trump slang} & trump & donald & donald\_trump & maga & high\_energy & centipede &  america\_great & meme \\ \toprule
     \textbf{The\_Donald ($t_1$)} & \textbf{immigration} & wall & illegal & border & immigrant & build\_wall & census & illegal\_alien & illegal\_immigration \\ \toprule
     \textbf{The\_Donald ($t_1$)} & \textbf{Mueller report} & mueller &  kavanaugh & trump\_supporter & office & hillary & comey & barr & collusion \\ \toprule
     \textbf{conspiracy ($t_0$)}& \textbf{surveillance} & cia & nsa & secret & operation & agency & agent & intelligence & surveillance \\ \toprule
     \textbf{conspiracy ($t_0$)}& \textbf{mass shootings} & shooting & police & shoot & shooter & suspect & sandy\_hook & kill & paris\_attack \\ \toprule
     \textbf{conspiracy ($t_1$)}& \textbf{Epstein} & epstein & jeffrey\_epstein & pedophile & ring & claim & pedophilia & pedo & dr \\ \toprule
     \textbf{conspiracy ($t_1$)}& \textbf{aliens} & alien & moon & earth & nasa & ufo & space & planet & sun \\ \toprule
    
    \bottomrule
    \end{tabular}
    }
\caption{A selection of clusters produced by $\wmdec$. Clustering facilitates inductive thematic discovery by grouping semantically similar words.}
\label{table:cluster_jack}
\end{table*}

We can also note that, though the top word distances are higher for \verb|r/The_Donald| at each period, the difference greatly shrinks, reflecting both an attenuation of the ritualized, insider language associated with Trump's online following during the 2016 election cycle, as well as an increase in talk related to Trump on \verb|r/conspiracy|. While these results are perhaps unsurprising, they demonstrate the greater richness of $\wmdec$ for the comparative analysis of corpora by introducing word-level data. Clustering words similarly allows for the discovery of pervasive thematic differences between corpora. Table \ref{table:cluster_jack} displays some example clusters, separated by subreddit and time period.

\subsection{Inductive discovery of thematic assimilation over time}

\begin{table}[]
\centering
\begin{tabular}{l l l}
\toprule
    \textbf{Word} & \textbf{CC (\%)} & \textbf{SWiC}\\
    \midrule
    
    hoax & -99.2 & discredit, truth, bogus \\
    fraudulent & -98.5 & theft, conspire, scam \\
    brainwash & -95.3 & teacher, public\_school \\
    threat & -67.0 & mitigate, deter \\
    jews & -60.4 & jewish, zionist \\
    surveillance & -49.6 & nsa, agency, agent \\
    alex\_jones & -39.0 & infowars, interview \\
    propaganda & -33.8 & false\_flag, agenda \\
    reality & -23.9 & veil, perceive \\

    \bottomrule
    \end{tabular}
\caption{CC (\%) = Cost change from $t_0$ to $t_1$, represented as a percentage, SWiC = Similar words in cluster (i.e., words from the same cluster whose cost also decreased from $t_0$ to $t_1$, while remaining in use in both subreddits).}
\label{table:words_w_negative_change}
\vspace{-15pt}
\end{table}

While $\wmdec$ can be used to identify distinguishing keywords, it can also be used to uncover fine-grained semantic assimilation. If a word $w$ strongly distinguishes corpus $S^a$ from corpus $S^b$ at $t_0$, but no longer at $t_1$, it could represent assimilation ($S^b$ starts using $w$ and related words more frequently), but it could equally represent $S^a$ ceasing to use $w$ (as is often the case with words related to current events). 

As such, the change in the distance contribution of a word over time must be considered alongside the change in word frequency. We might thus conceive of each word distance change as occupying a position in three-dimensional space, with axes corresponding to 1) change in aggregate word cost, 2) change in frequency in $S^a$, and 3) change in frequency in $S^b$. The same conceptualization can be applied to word clusters as well, wherein a cluster's values are simply the sum of its words' values.

Given our interest in semantic convergence, cases in which word distance contributions go down because usage ceases are of little interest (though they might interest others). Rather, we are interested in words which contribute less distance, while exhibiting comparable or increased use in both subreddits. Examining words meeting these criteria, and excluding low-frequency outliers, we indeed see many conspiratorial words whose cumulative distance contribution shrank from $t_0$ to $t_1$. A selection of such words is presented in Table \ref{table:words_w_negative_change}. These include terms such as \textit{hoax, fraudulent, brainwash, threat, jews, surveillance, alex\_jones, propaganda,} and \textit{reality}, to name a few. We can then examine other words in these keywords' clusters to get a sense of related words following similar patters of cost reduction and continued usage, as are displayed in the table. 

Of course, such observations are of exploratory nature, intended to demonstrate an inductive starting point for a more rigorous analysis, either within or outside the $\wmdec$ framework. We do not posit this as statistical proof of any process of assimilation. However, this general framework of comparative and longitudinal analysis allows for a much richer engagement with discursive change due to the robust semantic relationships captured by $\wmdec$ while still providing simple outputs, e.g., for significance testing. Despite its cursory nature, we hope this short vignette has demonstrated the interpretable potential of $\wmdec$, and how it might aid social science and humanities researchers with qualitative thematic discovery in large corpora, while also providing a computable metric for quantitative models.

\section{Conclusion}

We have introduced a novel iteration of $\wmd$, one which departs from the $\wmd$ variations which, despite their high performance on a variety of tasks, do away with one of the key contributions of the original model: the inherent and decomposable relationship between document-level semantic distances and the lexical semantics that underlie them. In so doing, we heed a recent, but hopefully long-lived, call for ML and particularly NLP researchers to prioritize models which are interpretable, providing explanatory value at the level of social meaning. We hope that our Python package might aid future researchers similarly interested in interpretable computational approaches. Furthermore, through the Gale-Shapley algorithm, we propose an approach for combining interpretability with environmental sustainability. Down the line, this framework could also be expanded to support dynamic embeddings from models such as BERT, which due to their technical opacity have yet to be widely adapted in sociocultural analysis.

Our vignette presented how one might use $\wmdec$ in a comparative and exploratory context. That is, we imagine this as somewhat of an inductive starting point---in this case, for an analysis of the relationship between Trumpist and conspiracy theorist online communities---revealing trends related to one's particular research question that can be pursued with more focused and rigorous analysis, be it qualitative, discourse analytic, or statistical. However, because of the rich, word-level quantitative measures $\wmdec$ provides, future research might attempt to employ it in, for example, causal models seeking to estimate the effect of a treatment condition on particular lexical features, time series analyses, and so on. 

\bibliography{emnlp2021}
\bibliographystyle{acl_natbib}

\appendix

\section{Text preprocessing and word2vec fine-tuning details}

The raw text data from the Pushshift dataset was processed as follows, all conducted in Python. Posts were first lowercased, and URLs were removed via regular expressions; stopwords were removed and remaining words were lemmatized, both using spaCy \cite{honnibal_spacy_2020}; markdown text and other special characters were removed using custom regex functions; finally, words occurring fewer than 20 times in the corpus (i.e., less than once every 1,000 documents) were removed. 

Once processed, the text was embedded with a word2vec model that was fine-tuned on a corpus of 533K self posts from these subreddits as well as other conspiracy and conservative subreddits, with preprocessing following the same steps as above. Fine-tuning was done using the skipgram implementation of word2vec, with negative sampling, a context window of 10 tokens, over four epochs and with a learning rate of 0.01. The model was phrased in a similar manner to the recommendations in \citet{mikolov_efficient_2013}, such that frequently co-occurring bigrams and trigrams were encoded as single lexical entities (e.g., \textit{president\_trump}). Further, each document was represented using Tf-Idf and L1-normalization, in order to prevent frequent words from crowding out information from less common but more salient words.

\begin{table}[!t]
\centering
\begin{tabular}{l l l l}
\toprule
\begin{tabular}{l c}
    \multicolumn{2}{c}{Positive} \\
    \textbf{Word} & \textbf{Cost}\\
    \midrule
    great & 41.79 \\
    best & 33.15 \\
    amazing & 31.35 \\ 
    friendly & 27.02 \\ 
    delicious & 26.18 \\ 
    highly\_recommend & 24.93 \\ 
    staff & 19.96 \\ 
    definitely & 19.52 \\ 
    always & 18.68 \\ 
    favorite & 18.42 \\ 
    massage  & 18.29 \\ 
    portland  & 16.15 \\ 
    \end{tabular}
    \begin{tabular}{l c}
    \multicolumn{2}{c}{Negative} \\
    \textbf{Word} & \textbf{Cost}\\
    \midrule
    told & 19.64\\ 
    order & 19.58\\ 
    rude  & 18.10\\ 
    never  & 17.94\\ 
    worst  & 17.44\\ 
    minute  & 15.14\\ 
    hour  & 14.97\\ 
    said  & 14.91\\ 
    money  & 13.89\\ 
    terrible  & 13.57\\ 
    called & 13.14\\
    asked & 13.05\\
    \end{tabular}
\\
\bottomrule
\end{tabular}
\caption{Decomposed word-level $\wmd_w$ distances for moving from the full set of positive reviews (left) the the full set of negative reviews and vice versa (right). Only the 12 words $w$ that contributed the most are shown.}
\label{tab:topn_yelp}
\end{table}

\begin{table*}[t!]
    \centering
    \resizebox{\textwidth}{!}{%
    \begin{tabular}{l c c c c c c c c c c}
    \toprule \\
    & \multicolumn{10}{c}{Positive to Negative Clusters} \\
      & 56 & 51 & 92 & 76 & 26 & 30 & 33 & 78 & 47 & 88\\
    \midrule
     Word (distance) & great (30.91) & made (3.18) & good (7.14) & delicious(21.13) & spot (8.24) &  friendly (20.8) 	&professional (9.19) & year (4.98) & massage (16.45) & quality (2.12) \\
     &best (26.33) & felt (2.85) & really (5.9) & favorite (15.06) & place (6.2) & staff (14.39) & team (6.35) & new (2.89) & relaxing (3.83) & price (1.97)
    \\ & amazing (25.21)  & soon (2.54) & nice (5.87) & loved (7.27) & town (5.07)  & everyone (6.24)  & thorough (3.54) & moved (2.77) & facial (3.72) & worth (1.95) \\
     & love (21.88) & quickly (2.07) & well (4.3) & fresh (7.24) & found (3.77) & attentive (5.6) & feel\_comfortable (3.48) & weekly (1.58) & spa (3.48) & option (1.94) \\
     & highly\_recommend (20.17) & hand (1.87) & kind (4.27) & tasty (4.78) & glad (3.36) & welcoming (4.06) & job (2.84) & st (1.03) & treatment (3.15) & beat (1.74) \\
    & definitely (15.54) & house (1.72) & lot (3.38) & try (4.7) & looking (3.0) & accommodating (3.77) & work (2.74) & year\_ago (0.98) & brow (2.81) & every\_penny (1.21) \\ & always (15.27) & next (1.36) & little (3.34) & enjoyed (4.43) & visit (2.95) & pleasant (3.28) & care (2.61) & first (0.86) & gentle (2.47) & size (0.95) \\
    & excellent (12.16)& around (1.22)& also (2.79)& must (4.39)& friend (2.62) & personable (2.76)& efficient (2.27)& monthly (0.73) & calming (2.32)& deal (0.95)\\
     & thank (10.63) & along (1.04) & enough (2.29) & die (2.95) & gem (2.52) & smile (2.38) & felt\_comfortable (2.19) & yr (0.71) & hot\_stone (2.2) & plus (0.9)\\
      & super (10.52) & totally (1.04) & much (2.0) & special (2.53) & coming (2.21) & met (1.46) & grateful (2.08) & last\_week (0.59) & lash (1.57) & penny (0.88) \\
    \bottomrule \\
    & \multicolumn{10}{c}{Negative to Positive Clusters} \\
      & 51 & 92 & 44 & 95 & 35 & 52 & 56 & 26 & 78 & 0
      \\
    \midrule
     Word (distance) & never (15.84) & like (4.84) & order (15.17) & worst (14.89) & told (16.11) & use (3.06) & ever (2.14) & review (5.09) & went (5.2) & terrible (11.24) \\
      & hour (11.77) & ok (3.46) & minute (12.0) & money (11.35) & said (11.41) & issue (2.41) & treat (1.23) & disappointed (3.2) & closed (4.6) & bad (7.88)\\
     & would (10.22) & hard (3.31) & waiting (7.83) & horrible (9.74) & asked (10.94) & follow (2.17) & unbelievable (0.98) & close (2.49) & month\_later (1.89) & awful (7.0) \\
     & u (7.87) & maybe (3.27) & table (7.56) & 	customer\_service (6.39) & manager (9.14) & problem (1.9) & cannot (0.63) & decided (1.6) & changed (1.87) & poor (6.84) \\
     & could (6.59) & however (2.43) & waited (6.24) & waste\_time (5.66) & refused (4.05) & frustrating (1.66) & allaround (0.28) & find (1.3) & update (1.85) & star (4.98)) \\
    & another (6.41) & used (2.41) & min (5.72) 	& please (4.4) & ask (2.86) & need (1.51) & every\_aspect (0.24) & read\_review (1.26) & day (1.8) & slow (4.86) \\
     & even (6.29) & seemed (2.33) & sitting (3.8) & unless (3.82) & asking (2.41) & properly (1.48) & everyday (0.2) & heard (1.24) & month\_ago (1.68) & worse (3.8)\\
     & nothing (6.06) & fine (2.31) & waiter (3.71) & somewhere\_else (3.63) & mistake (2.26) & multiple (1.27) & commend (0.2) & opened (0.82) & last (1.44) & disappointing (3.16) \\
    & give (5.36) & least (2.03) & hostess (3.59) & avoid (3.56) & stated (2.05) & due (1.17) & speedy (0.16) & somewhere (0.63) & twice (1.43) & sad (2.96) \\  &left (5.35) & something (1.71) & line (3.32) & suck (2.5) & informed (1.87) & failed (1.12) & always (0.13) & first\_impression (0.62) & month (1.23) & unfortunately (2.63) \\
    \bottomrule
    \end{tabular}
    }
\caption{The $\cmd_c$ for the top 10 $c$ when moving from the positive review set to the negative (top) and vice versa (bottom). The keywords and their order for each cluster are determined by the $\wmd_w$ of each word.}
\label{tab:clusters_yelp}
\end{table*}

\section{Sanity Check with Yelp Reviews}
\label{sec:appendix}

To further demonstrate the qualities of $\wmdec$, we offer an additional case study using a set of reviews from Yelp. This dataset is intuitive and well-known, and should hence complement our main analysis, as it requires less domain-knowledge for illustrating the basic qualities of $\wmdec$. To facilitate comparison with earlier work on the $\wmd$, we here rely on Euclidean distances, the metric which was used in the original $\wmd$ paper \cite{kusner_word_2015}, even though cosine distances are arguably preferable when working with semantic similarity tasks and therefore used in the main analysis of the paper. For the analysis in this appendix, we look at trends that are highly predicable, in order to provide a ``sanity check'' to ensure that our model was behaving as expected. Consequently, we looked at the $\wmd_w$ and $\mathrm{CMD}_c$ from positive and negative reviews to see whether the distance would, as expected, be composed mainly of different polarized sentiment words, such as \textit{good} and \textit{bad}. 

For this analysis, we use the latest version of the Yelp review dataset, accessed in late July 2021\footnote{https://www.yelp.com/dataset/}. The data was filtered to only include reviews written in the cities of Atlanta and Portland, two geographically and demographically distinct cities, after December 2017. We further filtered the data to only include reviews for establishments that were labeled with the categories  \textit{Restaurant} or \textit{Health \& Medical}. This way we wanted to show that category-specific trends would also emerge among positive and negative keywords, as well as explore whether any trends specific for the two cities would appear in the results. Reviews were then further filtered so that only highly positive (5 stars) or highly negative (1 star) were included. These were labeled positive and negative, respectively. Reviews were sampled from this pool, so that 2000 reviews were selected from both Atlanta and Portland, 500 negative and 500 positive for both restaurants and health services, for a total of 4000 reviews.

\subsection{Preprocessing and word vectors}
For the purposes of this analysis, we fine-tuned the same pretrained word2vec model that was used for the main analysis, with the same hyperparamters, but using the filtered Yelp dataset. A series of preprocessing steps were also conducted, including removal of URLs and stopwords and phrasing as proposed by \citet{mikolov_efficient_2013}. 

\subsection{Clustering and Gale-Shapley}
After preprocessing, clusters were detected in the word vectors using K-means. Through inspecting Silhouette scores and elbow plots, the number of clusters was chosen to be 100. Further, instead of performing the analysis on the full set of pairs, we first ran $\lcrwmd$ to find the $\rwmd$ between all pairs and then the Gale-Shapley (GS) algorithm to find the set of optimal pairs.

\subsection{Results from WMDecompose}

Next, the pairs from the GS matching were analysed using $\wmdec$. The words that added the most distance (with difference) when moving from the positive set to the negative and vice versa can be seen in Table \ref{tab:topn_yelp}. As expected, the top set of $\wmd_w$ word-level distances for the positive documents is dominated by positive sentiment words, with \textit{great}, \textit{best} and \textit{amazing} on top. However, other trends also appear, such as words specific to restaurants (\textit{delicious}) or health services (\textit{massage}). Interestingly enough, the word \textit{portland} appears as a high expense when moving from the positive to the negative set, indicating that positive reviewers located in that city might be more likely to mention it by name.

On the side of $\wmd_w$ word-level distances, the top words include negative sentiment words such as \textit{rude}, \textit{worst} and \textit{terrible} when moving from the negative to the positive reviews, but other patterns are also visible. Words about time, such as \textit{hour}, \textit{minute} and the word \textit{money}, contribute heavily to the semantic distance from negative to positive reviews, as do verbs often associated with verbal commands, such as \textit{told} and \textit{said}. Here, words related to the cities or categories of the reviews do not make the top 12 cut, although category specific words are present when looking at the top 50 (including \textit{insurance}, \textit{manager} and \textit{doctor}).

Switching focus to the $\cmd_c$ clusters displayed in Table \ref{tab:clusters_yelp}, we see some of the top words from Table \ref{tab:topn_yelp}, now organized by cluster. While there is some overlap in clusters when moving from positive to negative and vice versa (clusters  51, 92, 56, 26, 78), the keywords that define the clusters are different, as they are determined using the top ten $\wmd_w$ values of each word in the cluster. In Table \ref{tab:clusters_yelp}, we also see how category specific words, such as those in cluster 47 related to spa and massage services, add a significant distance when moving from positive to negative reviews.

\end{document}